\documentclass{article}
\usepackage[preprint]{colm2026_conference}

\usepackage{microtype}
\usepackage{hyperref}
\usepackage{url}
\usepackage{booktabs}
\usepackage{amsmath}
\usepackage{amssymb}
\usepackage{graphicx}
\usepackage{multirow}
\usepackage{array}
\usepackage{xcolor}
\usepackage{subcaption}
\usepackage{lineno}
\usepackage{colortbl}
\usepackage{float}
\usepackage{tcolorbox}
\tcbuselibrary{skins,breakable}
\usepackage{textcomp}  % suppresses OMS font substitution warnings

\definecolor{darkblue}{rgb}{0, 0, 0.5}

\definecolor{pairblue}{RGB}{219,234,254}   % BLUE_L  – pairwise path
\definecolor{pairbluedark}{RGB}{29,78,216} % BLUE_D
\definecolor{scoramber}{RGB}{254,243,199}  % AMB_L   – scoring path
\definecolor{scoramberdark}{RGB}{180,83,9} % AMB_D

\tcbset{
  promptbase/.style={
    enhanced, breakable,
    boxrule=1.2pt, arc=4pt,
    left=6pt, right=6pt, top=5pt, bottom=5pt,
    fontupper=\small\ttfamily,
    coltext=black,
    before skip=6pt,
    after skip=4pt,
  }
}

\newtcolorbox{pairbox}{
  promptbase,
  colback=pairblue,
  colframe=pairbluedark,
  title={\small\bfseries\color{white} Pairwise Preference Prompt},
  attach boxed title to top left={yshift=-2mm, xshift=4mm},
  boxed title style={
    colback=pairbluedark, colframe=pairbluedark,
    arc=3pt, boxrule=0pt
  },
}

\newtcolorbox{scorebox}{
  promptbase,
  colback=scoramber,
  colframe=scoramberdark,
  title={\small\bfseries\color{white} Direct Scoring Prompt},
  attach boxed title to top left={yshift=-2mm, xshift=4mm},
  boxed title style={
    colback=scoramberdark, colframe=scoramberdark,
    arc=3pt, boxrule=0pt
  },
}
\hypersetup{colorlinks=true, citecolor=darkblue, linkcolor=darkblue, urlcolor=darkblue}

\definecolor{lightgreen}{RGB}{230,255,230}
\definecolor{lightred}{RGB}{255,230,230}
\definecolor{lightyellow}{RGB}{255,255,220}

\title{Diagnosing LLM Judge Reliability:\\
Conformal Prediction Sets and Transitivity Violations}

\author{
  Manan Gupta\textsuperscript{1} \quad
  Dhruv Kumar\textsuperscript{1} \\
  \\
  \textsuperscript{1}BITS Pilani, Pilani Campus, India \\
  \texttt{\{f20241231, dhruv.kumar\}@pilani.bits-pilani.ac.in}
}

\newcommand{\judge}[1]{\textsc{#1}}
\newcommand{\crit}[1]{\textit{#1}}

\begin{document}

\ifcolmsubmission
\linenumbers
\fi

\maketitle

%% ================================================================
%% ABSTRACT
%% ================================================================
\begin{abstract}
LLM-as-judge frameworks are increasingly used for automatic NLG evaluation,
yet their per-instance reliability remains poorly understood.
We present a two-pronged diagnostic toolkit applied to SummEval:
\textbf{(1)} a transitivity analysis that reveals widespread per-input
inconsistency masked by low aggregate violation rates
($\bar{\rho} = 0.8$--$4.1\%$), with $33$--$67\%$ of documents
exhibiting at least one directed 3-cycle; and
\textbf{(2)} split conformal prediction sets over 1--5 Likert scores
providing theoretically-guaranteed $\geq(1{-}\alpha)$ coverage,
with set width serving as a per-instance reliability indicator
($r_s = {+}0.576$, $N{=}1{,}918$, $p < 10^{-100}$, pooled across all judges).
Critically, prediction set width shows consistent cross-judge agreement
($\bar{r} = 0.32$--$0.38$), demonstrating it captures
\emph{document-level difficulty} rather than judge-specific noise.
Across four judges and four criteria, both diagnostics converge:
\emph{criterion matters more than judge},
with \crit{relevance} judged most reliably
(avg.\ set size $\approx 3.0$) and \crit{coherence} moderately so
(avg.\ set size $\approx 3.9$), while \crit{fluency} and \crit{consistency}
remain unreliable (avg.\ set size $\approx 4.9$).
We release all code, prompts, and cached results.
\end{abstract}

%% ================================================================
%% 1. INTRODUCTION
%% ================================================================
\section{Introduction}
\label{sec:intro}

Automatic evaluation of natural language generation (NLG) has become
a cornerstone of modern NLP research.
LLM-as-judge systems, where a large language model scores or ranks
system outputs, have gained rapid adoption as scalable proxies for
human annotation~\citep{zheng2023judging,liu2023geval,fu2023gptscore}.
A single LLM call can replace expensive crowd-sourced annotation
pipelines, and practitioners increasingly treat these scores as
ground truth.

Yet a critical question goes largely unasked: \emph{when should you
trust an LLM judge?}
Aggregate metrics such as system-level Kendall's $\tau$ or Pearson
correlation with human scores look impressive, but they average over
hundreds of instances.
A judge that is right 90\% of the time can be spectacularly wrong on
the 10\% that matters most.
\textbf{This paper develops and evaluates two complementary diagnostics
for per-instance reliability.}

\begin{figure*}[t]
  \centering
  \includegraphics[width=\textwidth]{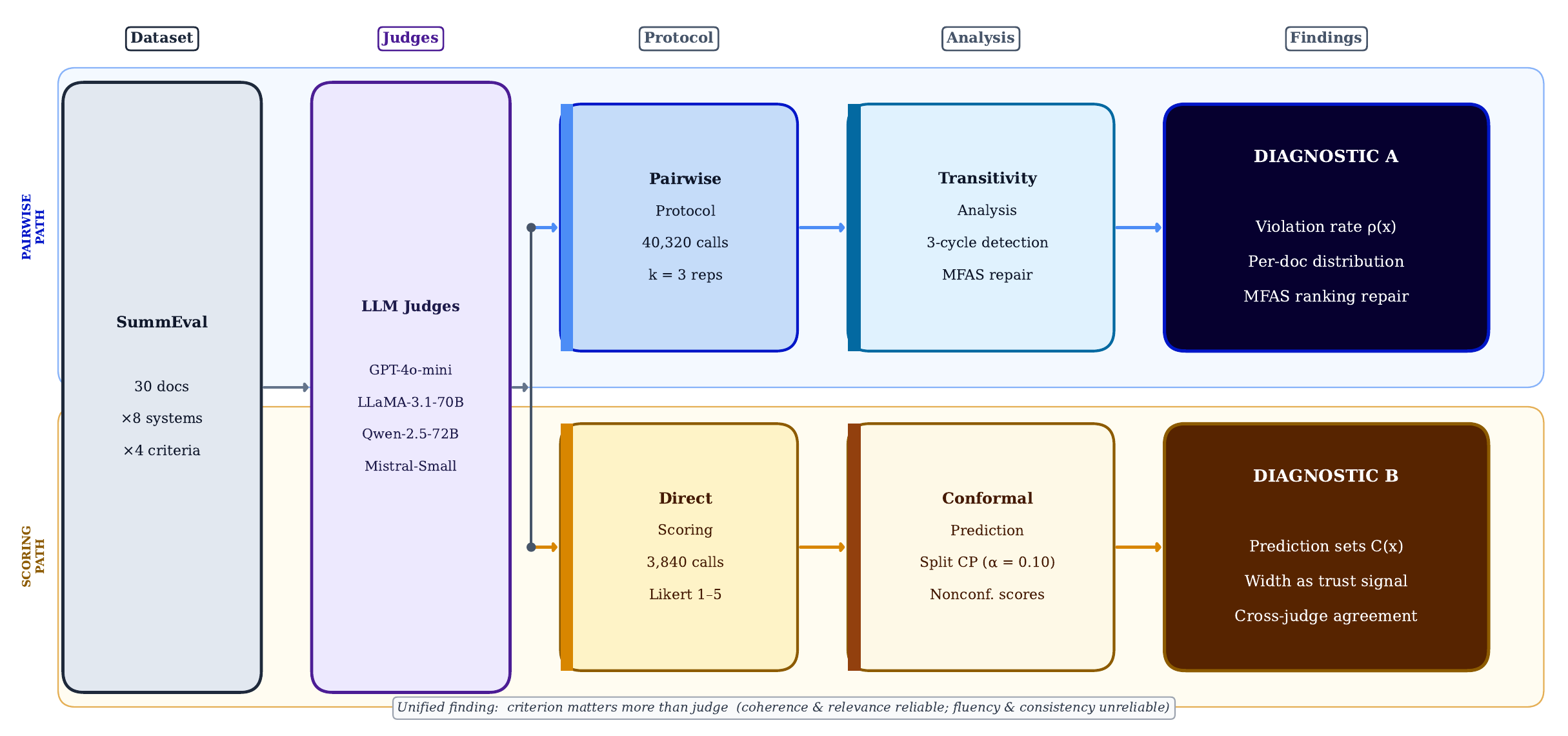}
  \caption{%
    \textbf{Two-pronged diagnostic pipeline.}
    SummEval documents are evaluated by four LLM judges under two protocols.
    The \emph{pairwise protocol} ($40{,}320$ API calls) feeds the
    transitivity diagnostic, which measures directed 3-cycle violation rates
    $\rho(x)$ per input and tests whether MFAS ranking repair improves
    agreement with human rankings.
    The \emph{direct scoring protocol} ($3{,}840$ API calls) feeds
    the conformal diagnostic, which produces prediction sets
    $\mathcal{C}(x)$ with guaranteed $\geq(1{-}\alpha)$ coverage;
    set width serves as a per-instance trust signal.}
  \label{fig:pipeline}
\end{figure*}

\paragraph{Contribution 1: Transitivity diagnostic.}
Pairwise LLM judges are assumed to produce transitive preferences,
yet tournament theory tells us that directed cycles
(A$\succ$B, B$\succ$C, C$\succ$A) arise naturally when alternatives
are near-equally preferred~\citep{young1988condorcet}.
We measure directed 3-cycle violation rates across four judges on
SummEval~\citep{fabbri2021summeval} and, crucially,
\emph{disaggregate by input document}.
Aggregate rates are low ($\bar{\rho} < 5\%$), but
$33$--$67\%$ of documents exhibit at least one violation,
with per-document rates reaching $30.4\%$ for Mistral-Small-3.1.
We also test whether Minimum Feedback Arc Set (MFAS) ranking
repair~\citep{ailon2008aggregating} improves agreement with human
rankings.It does not, establishing that violations are sparse noise
rather than systematic directional bias.

\paragraph{Contribution 2: Conformal prediction diagnostic.}
We apply split conformal prediction~\citep{vovk2005algorithmic,
angelopoulos2021gentle} to direct Likert scores,
producing prediction sets with finite-sample, distribution-free
coverage guarantees.
The central finding is that prediction set \emph{width} is a
statistically robust per-instance reliability indicator:
pooling $1{,}918$ observations across all judges and criteria yields
Spearman $r_s = {+}0.576$ ($p < 10^{-100}$) between width and
actual judge--human disagreement.
Moreover, different judges assign wide sets to the \emph{same documents}
($\bar{r} = 0.32$--$0.38$ across judge pairs for fluency, consistency,
and relevance), demonstrating that width measures inherent document
difficulty rather than judge-specific noise.

\paragraph{Unified finding.}
Both diagnostics independently identify the same axis of variation:
\crit{criterion} explains reliability more than \crit{judge}.
This is actionable: a practitioner deploying LLM judges should
trust coherence and relevance scores more than fluency and consistency
scores, regardless of which judge is used.

%% ================================================================
%% 2. RELATED WORK
%% ================================================================
\section{Related Work}
\label{sec:related}

\paragraph{LLM-as-judge reliability.}
\citet{zheng2023judging} introduced MT-Bench and Chatbot Arena,
establishing LLM judges as scalable evaluation tools.
\citet{liu2023geval} showed that GPT-4-based G-Eval correlates
strongly with human judgment on SummEval.
Known biases include position~\citep{wang2023large},
verbosity~\citep{saito2023verbosity}, and self-enhancement effects.
\citet{fernandes2023devil} and \citet{koo2023benchmarking} audit
LLM judges at scale, finding systematic weaknesses on specific
input types, consistent with our per-document view.
Concurrent to our work, \citet{ye2024flask} study fine-grained
reliability of LLM judges across skill categories, but without
formal uncertainty guarantees.

\paragraph{Transitivity and ranking consistency.}
Condorcet cycles in pairwise preferences have been studied in
social choice theory since~\citet{condorcet1785essay} and
are known to be ubiquitous when alternatives are
near-equal~\citep{young1988condorcet,moon1968topics}.
MFAS-based ranking repair has been applied to preference
aggregation~\citep{ailon2008aggregating} and recently to
LLM-generated ranked lists~\citep{qin2024large}.
We are the first to measure directed 3-cycle rates in LLM judges
at the per-document level and connect them to conformal uncertainty.

\paragraph{Conformal prediction in NLP.}
Split conformal prediction~\citep{vovk2005algorithmic,
papadopoulos2002inductive} provides distribution-free coverage
guarantees; \citet{angelopoulos2021gentle} give a modern treatment.
Applications to NLP include uncertainty for machine
translation~\citep{fomicheva2020unsupervised},
text classification~\citep{maltoudoglou2020bert},
and question answering~\citep{quach2024conformal}.
\citet{kumar2023conformal} apply conformal methods to LLM
generation quality, while \citet{kuhn2023semantic} propose
semantic entropy as a complementary uncertainty signal.
Our work is the first to apply conformal prediction to
\emph{LLM-as-judge scores} and to interpret prediction set width
as a per-instance deployment signal.

%% ================================================================
%% 3. METHODS
%% ================================================================
\section{Methods}
\label{sec:methods}

Figure~\ref{fig:pipeline} illustrates our pipeline.
Both diagnostics share the same four judges, dataset, and criteria,
enabling direct comparison of their findings.

\subsection{Transitivity Diagnostic}
\label{sec:methods:transitivity}

\paragraph{Tournament formulation.}
For each input document $x$ and set of $n$ system outputs
$\mathcal{S} = \{s_1,\ldots,s_n\}$, a pairwise judge produces a
tournament $G = (\mathcal{S}, E)$ where $(s_i, s_j) \in E$ iff
the judge prefers $s_i$ over $s_j$.
A \emph{transitivity violation} is a directed 3-cycle:
$s_i \succ s_j$, $s_j \succ s_k$, $s_k \succ s_i$.

\paragraph{Per-document violation rate.}
We define the violation rate for document $x$ as:
\begin{equation}
  \rho(x) = \frac{\#\,\text{directed 3-cycles in }G_x}{\binom{n}{3}},
\end{equation}
normalized by the total number of possible triples.
We report the aggregate mean $\bar{\rho}$, the fraction of documents
with $\rho(x) > 0$, and the full distribution $\{\rho(x)\}$.

\paragraph{Ranking methods.}
We compare five ranking methods:
\textbf{Win Rate} (fraction of pairwise wins),
\textbf{Bradley-Terry}~\citep{bradley1952rank} (maximum-likelihood
strength scores),
\textbf{Schulze}~\citep{schulze2011new} (beatpath method),
\textbf{MFAS-ILP} (exact Minimum Feedback Arc Set via integer
linear programming), and
\textbf{MFAS-Copeland} (Copeland scores as a fast MFAS approximation).
We measure agreement with human rankings via Kendall's $\tau$.

\paragraph{Repetitions.}
Each pair is queried $k{=}3$ times per judge to measure
win-rate confidence (0, 1/3, 2/3, or 1).

\subsection{Conformal Prediction Diagnostic}
\label{sec:methods:conformal}

\paragraph{Setup.}
We use split conformal prediction~\citep{vovk2005algorithmic} in
the \emph{direct scoring} setting: the judge assigns a Likert score
$\hat{y} \in \{1,\ldots,5\}$, and the calibration target is the
rounded average human score $y^* \in \{1,\ldots,5\}$.

\paragraph{Nonconformity score and prediction set.}
We use the absolute residual as the nonconformity score:
$s_i = |\hat{y}_i - y^*_i|$.
Given calibration set $\{(x_i, y^*_i)\}_{i=1}^n$, the conformal
threshold is:
\begin{equation}
  \hat{q} = s_{\left(\lceil(1-\alpha)(n+1)\rceil\right)},
\end{equation}
the appropriate empirical quantile ensuring marginal coverage
$\mathbb{P}(y^* \in \mathcal{C}(x)) \geq 1 - \alpha$~\citep{tibshirani2019conformal}.
The prediction set for a new instance with judge score $\hat{y}$ is:
\begin{equation}
  \mathcal{C}(x) = \bigl\{y \in \{1,\ldots,5\} : |\hat{y} - y| \leq \hat{q}\bigr\}.
\end{equation}
Set width $w(x) = |\mathcal{C}(x)|$ ranges from 1 (maximally confident)
to 5 (full uncertainty).

\paragraph{Evaluation protocol.}
We evaluate \textbf{(1)} empirical coverage vs.\ target $1{-}\alpha$;
\textbf{(2)} average set size as an informativeness measure;
\textbf{(3)} Spearman $r_s(w, |{\hat{y} - y^*}|)$ to quantify how
well set width predicts actual judge error; and
\textbf{(4)} inter-judge width agreement, the Spearman correlation
between widths assigned by two different judges to the same document,
which tests whether width reflects document-level difficulty
or judge-specific noise.
All metrics are averaged over 20 random 50/50 calibration/test splits
for stable estimates.

%% ================================================================
%% 4. EXPERIMENTAL SETUP
%% ================================================================
\section{Experimental Setup}
\label{sec:experiments}

\paragraph{Dataset.}
SummEval~\citep{fabbri2021summeval} contains 100 documents
$\times$ 16 systems ($=1{,}600$ outputs) with human Likert scores
(1--5) on \crit{coherence}, \crit{consistency}, \crit{fluency},
and \crit{relevance}, averaged over three annotators.
We subsample to 30 documents $\times$ 8 systems
(systems 0, 2, 4, 6, 9, 11, 13, 15) for cost efficiency,
rounding averaged human scores to the nearest integer for conformal
calibration.

\paragraph{Judges.}
We evaluate four instruction-tuned LLMs accessed via OpenRouter:
\judge{GPT-4o-mini} (\texttt{gpt-4o-mini}),
\judge{LLaMA-3.1-70B} (\texttt{meta-llama/llama-3.1-70b-instruct}),
\judge{Qwen-2.5-72B} (\texttt{qwen/qwen-2.5-72b-instruct}), and
\judge{Mistral-Small-3.1} (\texttt{mistralai/mistral-small-3.1-24b-instruct}).
All responses are cached in SQLite.

%% ================================================================
%% 5. RESULTS
%% ================================================================
\section{Results}
\label{sec:results}

\subsection{Transitivity Violations: Aggregate Rates Mask Per-Document Heterogeneity}
\label{sec:results:transitivity}

\begin{table}[t]
\centering
\small
\setlength{\tabcolsep}{5pt}
\begin{tabular}{lcccc}
\toprule
\textbf{Judge} & \textbf{Agg.\ $\bar{\rho}$} & \textbf{\%\,docs $\geq$1} & \textbf{Max $\rho$} & \textbf{Med.\ $\rho$} \\
\midrule
\judge{LLaMA-3.1-70B}   & 0.008 & 33.3\% &  3.6\% & 0.0\% \\
\judge{Qwen-2.5-72B}    & 0.022 & 50.0\% & 14.3\% & 1.8\% \\
\judge{Mistral-Small}   & 0.041 & 50.0\% & \textbf{30.4\%} & 1.8\% \\
\judge{GPT-4o-mini}     & 0.014 & 46.7\% &  7.1\% & 0.0\% \\
\bottomrule
\end{tabular}
\caption{Per-document transitivity violation statistics (\crit{coherence}).
Aggregate rates $\bar{\rho}$ appear low (${<}5\%$), yet $33$--$50\%$ of
documents exhibit at least one directed 3-cycle, and per-document rates
reach $30.4\%$ for \judge{Mistral-Small}.
Median = 0 for all judges: most documents are violation-free, but a
minority expose severe judge inconsistency.}
\label{tab:violations}
\end{table}

Table~\ref{tab:violations} reveals a fundamental measurement problem.
Aggregate violation rates $\bar{\rho}$ range from $0.8\%$ to $4.1\%$
across judges, statistics that would reassure any practitioner.
But the per-document view tells a different story:
between $33\%$ and $50\%$ of documents carry at least one directed 3-cycle,
and the worst document for \judge{Mistral-Small} has $\rho(x) = 30.4\%$,
meaning nearly a third of all triples form preference cycles on that input.

\begin{figure}[t]
  \centering
  \includegraphics[width=\linewidth]{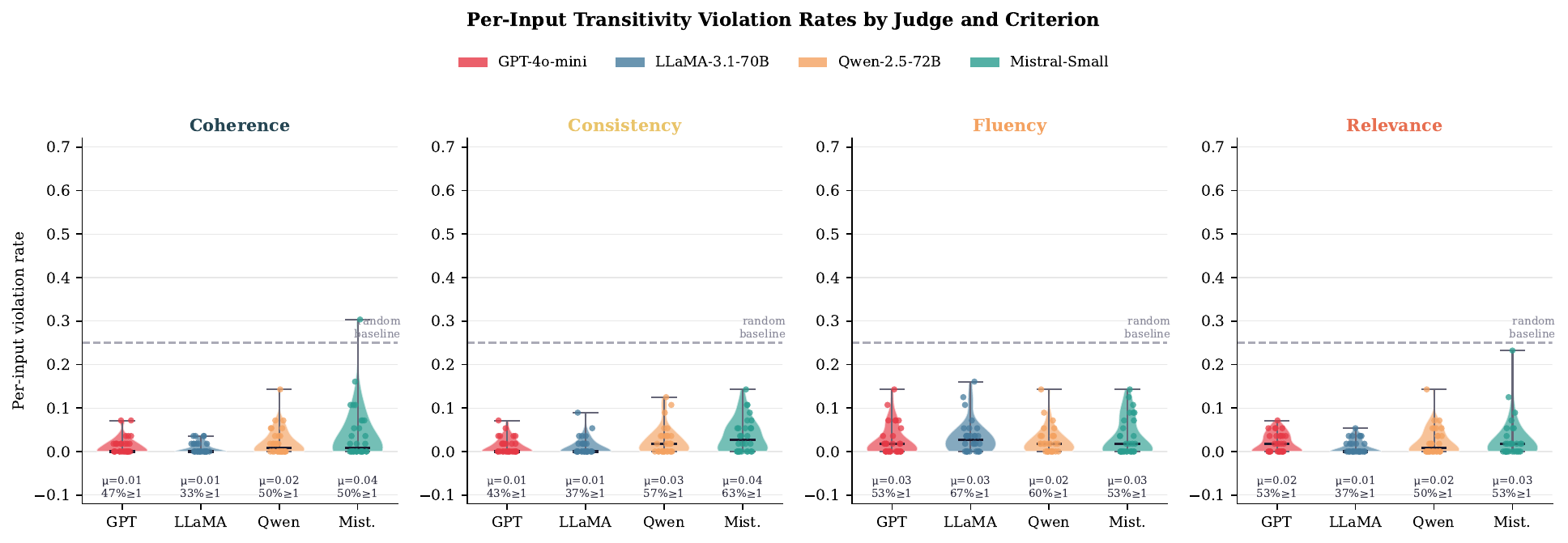}
  \caption{%
    \textbf{Per-document violation rate distributions.}
    Each violin shows the distribution of $\rho(x)$ across 30 documents
    for one judge--criterion pair.
    Dashed horizontal line: random-baseline rate (0.25).
    All distributions are right-tailed with median $= 0$,
    but the upper tails, where a single document can expose
    $>30\%$ violation rates, are practically significant.
    \crit{Fluency} consistently shows the widest tails.}
  \label{fig:violations}
\end{figure}

This is not random noise.
Figure~\ref{fig:violations} shows that the per-document distribution is
heavily right-tailed: the median is zero across all judges, but a small
fraction of documents, typically those where system outputs differ
subtly in quality, drive the aggregate statistic.
A practitioner who relies on $\bar{\rho}$ alone will miss this
concentrated inconsistency.

Appendix~\ref{app:violations} (Table~\ref{tab:full_violations}) provides
violation statistics across all four criteria.
\crit{Fluency} and \crit{consistency} show the highest fraction
of documents with $\geq 1$ violation (up to 66.7\% for
LLaMA--fluency), converging with the conformal results below.

\paragraph{MFAS ranking repair does not help.}
Table~\ref{tab:mfas} compares five ranking methods against the
SummEval human gold standard on \crit{coherence}.
MFAS-ILP achieves the highest Kendall's $\tau$ on one judge
but matches or falls below Win Rate on the other three.
No method consistently dominates.
This non-result is informative: when violations are sparse and
concentrated in a few documents, the tournament graph is nearly
acyclic and MFAS repair finds little to improve.
\emph{The diagnostic value of transitivity analysis lies in
identifying unreliable documents, not in repairing global rankings.}

\begin{table}[t]
\centering
\small
\setlength{\tabcolsep}{4pt}
\begin{tabular}{lcccc}
\toprule
\textbf{Method} & \textbf{LLaMA} & \textbf{Qwen} & \textbf{Mistral} & \textbf{GPT} \\
\midrule
Win Rate             & 0.571 & 0.643 & 0.071 & \textbf{0.714} \\
Bradley-Terry        & 0.571 & 0.643 & 0.071 & \textbf{0.714} \\
Schulze              & \textbf{0.643} & 0.643 & 0.071 & 0.643 \\
MFAS-Copeland        & 0.571 & 0.643 & 0.071 & \textbf{0.714} \\
MFAS-ILP$^\dagger$   & \textbf{0.643} & \textbf{0.643} & \textbf{0.071} & 0.643 \\
\bottomrule
\end{tabular}
\vspace{2pt}
{\footnotesize $^\dagger$ Exact ILP solution; \textbf{bold} = best per judge.}
\caption{Ranking agreement (Kendall's $\tau$) vs.\ SummEval human
gold standard (\crit{coherence}).
MFAS does not consistently outperform Win Rate or Bradley-Terry.
When violations are sparse, standard aggregation already captures
the signal; MFAS repair adds noise rather than structure.}
\label{tab:mfas}
\end{table}

\subsection{Conformal Prediction Sets: Guaranteed Coverage with a Trust Signal}
\label{sec:results:conformal}

\begin{figure}[t]
  \centering
  \includegraphics[width=\linewidth]{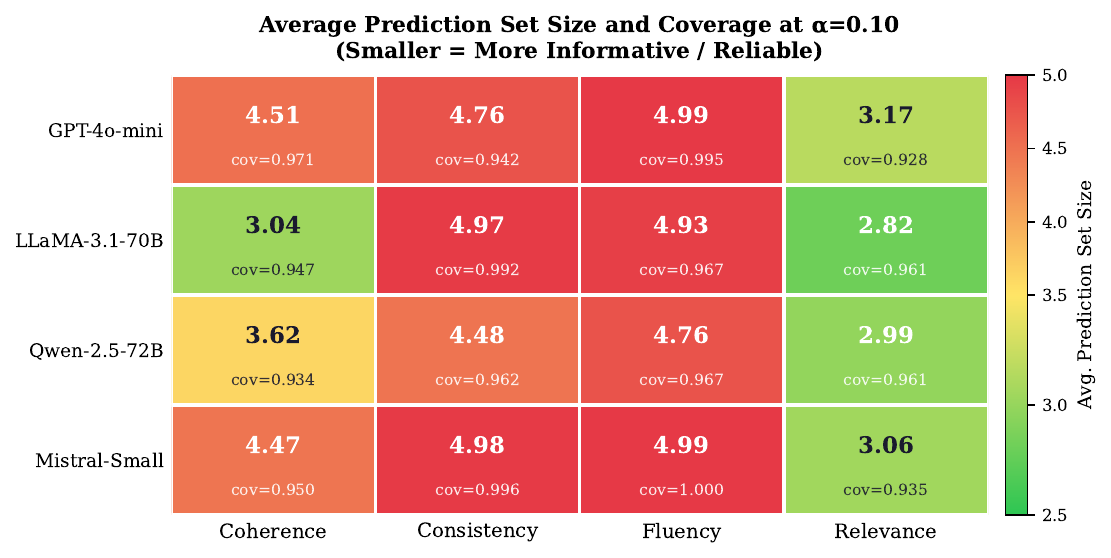}
  \caption{%
    \textbf{Average prediction set size at $\alpha{=}0.10$
    (green $=$ small $=$ reliable; red $=$ large $=$ unreliable).}
    Each cell shows average set size (larger text) and empirical
    coverage (smaller text).
    The criterion axis drives variation far more than the judge axis:
    \crit{coherence} and \crit{relevance} (left two columns) are
    reliably judged (${\approx}3.0$), while \crit{fluency} and
    \crit{consistency} are near-maximally uncertain (${\approx}5.0$).
    All 16 cells meet the 90\% coverage guarantee.}
  \label{fig:heatmap}
\end{figure}

\paragraph{Coverage guarantee satisfied.}
The conformal guarantee holds for \emph{all} 16
judge$\times$criterion combinations across all four $\alpha$ levels
tested ($\alpha \in \{0.05, 0.10, 0.15, 0.20\}$).
Figure~\ref{fig:coverage} confirms that empirical coverage tracks
the theoretical $1{-}\alpha$ line, meeting or exceeding it at every
operating point.
Full coverage and set-size results are in Appendix~\ref{app:full_conformal}.

\begin{table}[t]
\centering
\small
\setlength{\tabcolsep}{4pt}
\begin{tabular}{ll
  >{\columncolor{lightgreen}}c
  c
  >{\columncolor{lightred}}c
  >{\columncolor{lightred}}c
  c}
\toprule
\textbf{Judge} & & \textbf{Coh.} & \textbf{Rel.} & \textbf{Con.} & \textbf{Flu.} & \textbf{$r_s$\,(w,e)} \\
\midrule
\multirow{2}{*}{\judge{GPT-4o-mini}}
  & Size & 4.51 & 3.17 & 4.76 & 4.99 & \\
  & Cov. & .971 & .928 & .942 & .995 & $+$0.18\,/\,$+$0.25 \\[2pt]
\multirow{2}{*}{\judge{LLaMA-3.1-70B}}
  & Size & 3.04 & 2.82 & 4.97 & 4.93 & \\
  & Cov. & .947 & .961 & .992 & .967 & $-$0.03\,/\,$+$0.08 \\[2pt]
\multirow{2}{*}{\judge{Qwen-2.5-72B}}
  & Size & 3.62 & 2.99 & 4.48 & 4.76 & \\
  & Cov. & .934 & .961 & .962 & .967 & $+$0.14\,/\,$\mathbf{+0.65}$ \\[2pt]
\multirow{2}{*}{\judge{Mistral-Small}}
  & Size & 4.47 & 3.06 & 4.98 & 4.99 & \\
  & Cov. & .950 & .935 & .996 & 1.00 & $-$0.02\,/\,$+$0.17 \\
\bottomrule
\end{tabular}
\caption{Conformal results at $\alpha{=}0.10$ (90\% target coverage).
Columns ordered: \colorbox{lightgreen}{reliable criteria} then
\colorbox{lightred}{unreliable criteria}.
$r_s$\,(w,e) = Spearman correlation of set width with absolute error,
reported as Coh./Con.\ (coherence and consistency).
\judge{Qwen-2.5-72B} achieves the strongest width--error signal
for consistency ($r_s = +0.65$).}
\label{tab:conformal}
\end{table}

\paragraph{Criterion drives reliability, not judge.}
Table~\ref{tab:conformal} and Figure~\ref{fig:heatmap} show the
dominant pattern: the criterion column explains far more variance
in average set size than the judge row.
\crit{Coherence} and \crit{relevance} receive small sets (2.82--4.51
labels) while \crit{fluency} and \crit{consistency} receive near-maximal
sets (4.47--4.99) across all four judges.

This is not a failure of the method.
SummEval neural summaries are uniformly fluent, leaving little
variance for a judge to exploit~\citep{fabbri2021summeval}.
Consistency requires cross-document factual reasoning that 24--72B
scale models perform inconsistently.
The wide prediction sets are the method \emph{correctly reporting}
that it cannot reliably grade these criteria.

\paragraph{Set width predicts judge error.}
Figure~\ref{fig:reliability} shows pooled reliability diagrams for
each criterion: mean absolute error (MAE) vs.\ prediction set width,
pooled across all four judges.
Thirteen of sixteen judge$\times$criterion combinations show a perfectly
monotonic width--error relationship.
Pooling all 1,918 observations, Spearman $r_s = {+}0.576$
($p < 10^{-100}$).
Per-criterion pooled correlations are $r_s = {+}0.34$ (consistency),
$r_s = {+}0.16$ (coherence), $r_s = {+}0.15$ (fluency),
and $r_s = {-}0.01$ (relevance, n.s.).

The three non-monotonic cases, LLaMA fluency, LLaMA relevance, and
Mistral coherence, share a common structure: only two distinct width
values are observed (widths 2 and 3), leaving insufficient variance
for a meaningful per-judge correlation.
The width--error relationship is real but requires pooling to emerge
against this discrete noise floor.

\begin{figure}[t]
  \centering
  \includegraphics[width=\linewidth]{figures/fig3_reliability.pdf}
  \caption{%
    \textbf{Pooled reliability diagrams} (all four judges, $\alpha{=}0.10$).
    $x$-axis: prediction set width; $y$-axis: mean absolute error (MAE)
    vs.\ human score. Error bars: 95\% CI. Annotations: sample count per width.
    Spearman $r_s$ and $p$-value shown per panel.
    \crit{Consistency} shows the clearest signal ($r_s = {+}0.34$, $p{<}0.0001$);
    \crit{relevance} is the exception ($r_s \approx 0$, $p{=}0.86$).}
  \label{fig:reliability}
\end{figure}

\paragraph{Width reflects document difficulty, not judge noise.}
A potential confound: perhaps wide sets indicate that a specific judge
is unreliable, rather than that the document is inherently hard to
evaluate.
Table~\ref{tab:cross_judge} addresses this directly by measuring
inter-judge agreement on prediction set width: the Spearman correlation
between widths assigned by two \emph{different} judges to the same
documents.

For \crit{fluency}, \crit{consistency}, and \crit{relevance},
15 of 18 judge pairs show significant positive width agreement
($p < 0.05$), with mean correlations of $\bar{r} = 0.38$, $0.32$,
and $0.36$ respectively.
The standout pairs, GPT/Mistral fluency ($r = +0.81$) and
GPT/Qwen relevance ($r = +0.70$), confirm that different model
families converge on the same documents as hard to judge.
\crit{Coherence} is the exception ($\bar{r} = 0.10$), consistent with
its smaller and less variable set sizes.

\begin{table}[t]
\centering
\small
\setlength{\tabcolsep}{5pt}
\begin{tabular}{lcccc}
\toprule
\textbf{Judge Pair} & \textbf{Coh.} & \textbf{Con.} & \textbf{Flu.} & \textbf{Rel.} \\
\midrule
GPT / LLaMA    & $+$0.12      & $+$0.22\dag   & $+$0.45\ddag  & $+$0.28\ddag  \\
GPT / Qwen     & $+$0.27\ddag & $+$0.38\ddag  & $+$0.23\dag   & $\mathbf{+0.70}$\ddag \\
GPT / Mistral  & $+$0.31\ddag & $+$0.32\ddag  & $\mathbf{+0.81}$\ddag & $+$0.29\ddag  \\
LLaMA / Qwen   & $-$0.08      & $+$0.17       & $+$0.21\dag   & $+$0.20\dag   \\
LLaMA / Mistral& $+$0.08      & $\mathbf{+0.56}$\ddag & $+$0.23\dag   & $+$0.25\dag   \\
Qwen / Mistral & $-$0.09      & $+$0.24\ddag  & $+$0.33\ddag  & $+$0.44\ddag  \\
\midrule
\textit{Mean}  & \textit{$+$0.10} & \textit{$+$0.32} & \textit{$+$0.38} & \textit{$+$0.36} \\
\bottomrule
\multicolumn{5}{l}{\footnotesize \dag\,$p{<}.05$\quad \ddag\,$p{<}.01$ or better\quad \textbf{bold}\,$r{>}0.50$}
\end{tabular}
\caption{Inter-judge agreement on prediction set width (Spearman $r$,
$\alpha{=}0.10$). Positive values indicate different judges assign wider
sets to the same documents, width reflects \emph{document-level difficulty},
not judge-specific noise.}
\label{tab:cross_judge}
\end{table}

\begin{figure}[t]
  \centering
  \includegraphics[width=\linewidth]{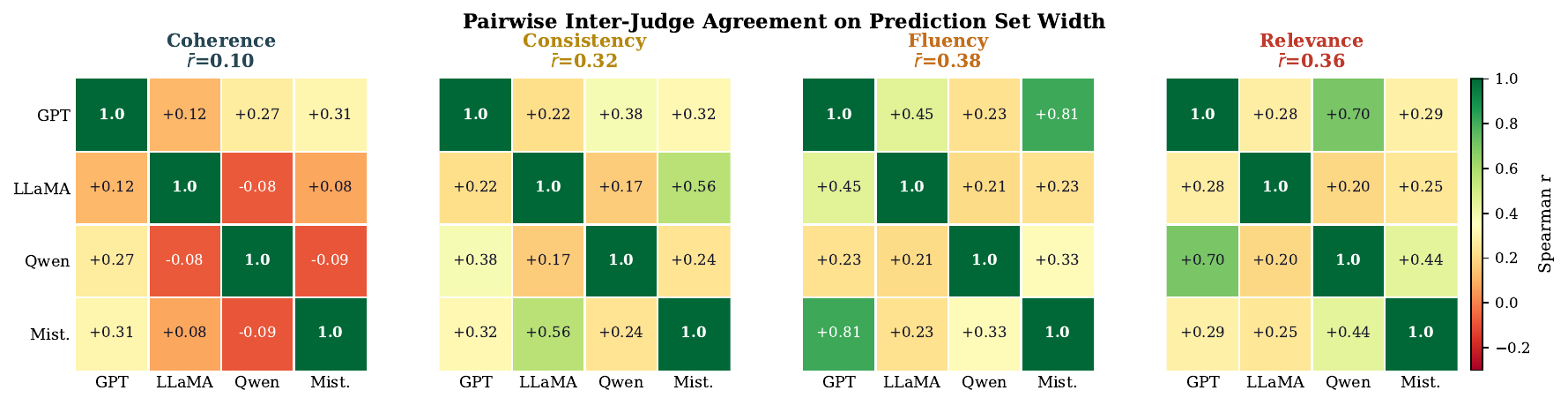}
  \caption{%
    \textbf{Inter-judge width agreement matrices} (Spearman $r$,
    $\alpha{=}0.10$). Rows/columns: the four judges.
    Diagonal forced to 1.0. \crit{Coherence} (leftmost) shows
    predominantly near-zero off-diagonal entries; \crit{fluency}
    and \crit{relevance} show consistently positive agreement,
    confirming that prediction width tracks document-level
    difficulty across model families.}
  \label{fig:agreement}
\end{figure}

Figure~\ref{fig:agreement} visualizes the full inter-judge agreement
matrices, making the qualitative pattern clear: coherence shows near-zero
off-diagonal entries while fluency and relevance show predominantly
warm (positive) off-diagonal colors.

\subsection{Convergent Evidence: Criterion Drives Reliability}
\label{sec:results:convergent}

Both diagnostics, applied independently, reach the same conclusion.
\crit{Fluency} and \crit{consistency} show the highest fraction of
documents with $\geq 1$ transitivity violation
(LLaMA: $66.7\%$ for fluency; Qwen: $60\%$; Table~\ref{tab:full_violations})
\emph{and} the widest prediction sets (avg.\ size $> 4.9$ for three of
four judges).
\crit{Coherence} and \crit{relevance} show lower violation rates
\emph{and} narrower prediction sets (avg.\ size $< 4.0$ for three
of four judges).

This cross-diagnostic agreement is strong evidence that criterion
difficulty is a fundamental property of the evaluation task,
not an artifact of any single measurement approach.
A document that causes preference cycles in pairwise ranking also
receives a wide prediction interval in direct scoring, both signals
point to the same underlying difficulty.

%% ================================================================
%% 6. DISCUSSION
%% ================================================================
\section{Discussion}
\label{sec:discussion}

\paragraph{The masked-heterogeneity problem.}
A practitioner who reports only $\bar{\rho} < 5\%$ or system-level
$\tau > 0.5$ will conclude that their LLM judge is reliable.
Our results show this conclusion is premature: nearly half of all
documents expose judge inconsistency at the per-document level.
We recommend that evaluation studies report, at minimum:
\textbf{(a)} the fraction of documents with $\geq 1$ violation;
and \textbf{(b)} the distribution of per-document violation rates,
not just the aggregate mean.

\paragraph{Prediction sets as a deployment signal.}
Our conformal results have an immediate practical implication:
before accepting an LLM judge score, compute the prediction set.
\begin{itemize}
  \item If $|\mathcal{C}(x)| \leq 2$: the judge is likely reliable
    for this instance, proceed.
  \item If $|\mathcal{C}(x)| = 5$ (full scale): the judge expresses
    maximum uncertainty, consider human annotation.
\end{itemize}
This selective escalation strategy is principled: the coverage guarantee
ensures $\mathcal{C}(x)$ contains the human score with at least
$1{-}\alpha$ probability.
The cross-judge agreement results further justify this approach:
a wide set from any judge is a warning about the document,
not about that specific model.

\paragraph{Why MFAS does not help.}
When violations are sparse ($\bar{\rho} < 5\%$) and concentrated in
a minority of documents, the tournament graph is nearly acyclic.
MFAS repair finds at most a handful of edges to reverse, insufficient
to change aggregate rankings.
Detecting these problematic documents for human follow-up is a more
effective use of the transitivity signal than attempting automated repair.

\paragraph{The coherence exception.}
\crit{Coherence} is the one criterion where inter-judge width
agreement is weak ($\bar{r} = 0.10$).
We hypothesize two reasons: (1) neural summaries in SummEval vary
substantially in coherence, making it a more discriminable dimension;
and (2) different model families may have different internal
representations of ``coherence,'' leading to idiosyncratic scoring
patterns that do not generalize across judges.

%% ================================================================
%% 7. LIMITATIONS
%% ================================================================
\section{Limitations}
\label{sec:limitations}

\textbf{Scale and generalization.}
We use 30 documents $\times$ 8 systems from SummEval.
Results may differ on larger subsets, other summarization datasets,
or non-summarization tasks (dialogue, translation, etc.).

\textbf{Marginal vs.\ conditional coverage.}
Split conformal guarantees \emph{marginal} coverage
$\mathbb{P}(y^* \in \mathcal{C}(x)) \geq 1{-}\alpha$, not
per-document conditional coverage.
Harder documents may receive tighter-than-justified sets in
practice; conditional conformal methods~\citep{angelopoulos2021gentle}
could address this.

\textbf{Fixed nonconformity score.}
We use the absolute residual $|\hat{y} - y^*|$.
Learned nonconformity scores (e.g., based on judge confidence or
LLM log-probabilities) could produce tighter, more informative sets.

\textbf{Prompt sensitivity.}
Each judge uses a single prompt template per criterion.
Different prompts may yield different violation rates and set widths;
we leave prompt-robustness analysis to future work.

\textbf{Human score rounding.}
SummEval provides averaged annotations; rounding to integers
introduces a small discretization error in calibration targets.

%% ================================================================
%% 8. CONCLUSION
%% ================================================================
\section{Conclusion}
\label{sec:conclusion}

We presented two complementary, low-cost diagnostics for LLM judge
reliability.
Transitivity analysis reveals that per-document inconsistency is
dramatically higher than aggregate statistics suggest, a pattern
invisible to standard evaluation metrics.
Conformal prediction sets provide finite-sample coverage guarantees
and a practical deployment signal: prediction set width predicts
actual judge error (pooled $r_s = {+}0.576$, $N{=}1{,}918$) and
tracks document-level difficulty rather than judge-specific noise.

Both diagnostics independently converge on the same finding:
\emph{criterion matters more than judge}.
Coherence and relevance can be judged reliably by any of the four
models tested; fluency and consistency should be treated with
skepticism regardless of the model.
We recommend that LLM evaluation pipelines adopt per-instance
uncertainty reporting as standard practice, and we release all
code, prompts, and cached API responses to support reproducibility.

%% ================================================================
%% ACKNOWLEDGMENTS
%% ================================================================

%% ================================================================
%% ETHICS STATEMENT
%% ================================================================
\section*{Ethics Statement}
This work develops diagnostic tools for assessing the reliability of
LLM-as-judge systems, with the goal of reducing uncritical reliance on
automated evaluation.
All experiments use publicly available data (SummEval) and commercially
available LLMs accessed via a standard API.
We do not collect human annotations, and the research poses no direct risk
of harm to individuals.
The broader impact is positive: surfacing systematic failure modes of LLM
judges helps practitioners deploy automated evaluation more responsibly and
identify when human oversight is needed.

%% ================================================================
%% REFERENCES
%% ================================================================
\bibliography{references}
\bibliographystyle{colm2026_conference}

%% ================================================================
%% APPENDIX
%% ================================================================
\appendix

\section{Prompt Templates}
\label{app:prompts}

\begin{pairbox}
You are evaluating the \{criterion\} of two summaries\\
of the following article.\\
\\
Article: \{document\}\\
\\
Summary A\@: \{system\_a\}\\
Summary B\@: \{system\_b\}\\
\\
Which summary is better in terms of \{criterion\}?\\
Answer with exactly ``A'' or ``B''. No explanation.
\end{pairbox}

\begin{scorebox}
You are evaluating the \{criterion\} of the following\\
summary of an article.\\
\\
Article: \{document\}\\
Summary: \{system\_output\}\\
\\
Rate the \{criterion\} on a scale of 1--5 where:\\
1 = Very Poor,\ \ 2 = Poor,\ \ 3 = Fair,\\
4 = Good,\ \ 5 = Excellent\\
\\
Respond with a single integer between 1 and 5.\\
No explanation needed.
\end{scorebox}

\vspace{-4pt}
\section{Full Conformal Results}
\label{app:full_conformal}

Table~\ref{tab:full_conformal} reports conformal coverage and average
set size at all four $\alpha$ levels tested.
Every entry exceeds the $1{-}\alpha$ target, confirming the theoretical
guarantee holds robustly across all operating points.

\begin{table}[H]
\centering
\caption{Full conformal prediction results across all $\alpha$ levels.
Mean coverage and set size across 20 random splits.
All coverages meet or exceed the $1{-}\alpha$ target (bolded).}
\label{tab:full_conformal}
\resizebox{\linewidth}{!}{%
\setlength{\tabcolsep}{4pt}
\begin{tabular}{llcccccccc}
\toprule
\multirow{2}{*}{Judge} & \multirow{2}{*}{Criterion}
  & \multicolumn{2}{c}{$\alpha=0.05$}
  & \multicolumn{2}{c}{$\alpha=0.10$}
  & \multicolumn{2}{c}{$\alpha=0.15$}
  & \multicolumn{2}{c}{$\alpha=0.20$} \\
\cmidrule(lr){3-4}\cmidrule(lr){5-6}\cmidrule(lr){7-8}\cmidrule(lr){9-10}
& & Cov. & Size & Cov. & Size & Cov. & Size & Cov. & Size \\
\midrule
\multirow{4}{*}{\judge{LLaMA-3.1-70B}}
  & Coherence   & \textbf{.993} & 4.29 & \textbf{.947} & 3.04 & \textbf{.945} & 2.98 & \textbf{.945} & 2.98 \\
  & Consistency & \textbf{.992} & 4.97 & \textbf{.992} & 4.97 & \textbf{.889} & 4.26 & \textbf{.889} & 4.26 \\
  & Fluency     & \textbf{.971} & 4.95 & \textbf{.967} & 4.93 & \textbf{.967} & 4.93 & \textbf{.967} & 4.93 \\
  & Relevance   & \textbf{.964} & 2.98 & \textbf{.961} & 2.82 & \textbf{.961} & 2.82 & \textbf{.961} & 2.82 \\
\midrule
\multirow{4}{*}{\judge{Qwen-2.5-72B}}
  & Coherence   & \textbf{1.000} & 4.55 & \textbf{.934} & 3.62 & \textbf{.905} & 2.99 & \textbf{.905} & 2.99 \\
  & Consistency & \textbf{.963} & 4.53 & \textbf{.962} & 4.48 & \textbf{.962} & 4.48 & \textbf{.962} & 4.48 \\
  & Fluency     & \textbf{.967} & 4.76 & \textbf{.967} & 4.76 & \textbf{.967} & 4.76 & \textbf{.967} & 4.76 \\
  & Relevance   & \textbf{.975} & 3.63 & \textbf{.961} & 2.99 & \textbf{.961} & 2.99 & \textbf{.961} & 2.99 \\
\midrule
\multirow{4}{*}{\judge{Mistral-Small}}
  & Coherence   & \textbf{.980} & 4.83 & \textbf{.950} & 4.47 & \textbf{.950} & 4.47 & \textbf{.901} & 4.10 \\
  & Consistency & \textbf{.996} & 4.98 & \textbf{.996} & 4.98 & \textbf{.996} & 4.98 & \textbf{.970} & 4.94 \\
  & Fluency     & \textbf{1.000} & 4.99 & \textbf{1.000} & 4.99 & \textbf{1.000} & 4.99 & \textbf{1.000} & 4.99 \\
  & Relevance   & \textbf{.995} & 4.15 & \textbf{.935} & 3.06 & \textbf{.931} & 2.93 & \textbf{.931} & 2.93 \\
\midrule
\multirow{4}{*}{\judge{GPT-4o-mini}}
  & Coherence   & \textbf{.979} & 4.59 & \textbf{.971} & 4.51 & \textbf{.899} & 3.46 & \textbf{.876} & 2.98 \\
  & Consistency & \textbf{.996} & 4.99 & \textbf{.942} & 4.76 & \textbf{.899} & 4.54 & \textbf{.899} & 4.54 \\
  & Fluency     & \textbf{.995} & 4.99 & \textbf{.995} & 4.99 & \textbf{.995} & 4.99 & \textbf{.984} & 4.97 \\
  & Relevance   & \textbf{1.000} & 4.37 & \textbf{.928} & 3.17 & \textbf{.919} & 2.96 & \textbf{.919} & 2.96 \\
\bottomrule
\end{tabular}
}% end resizebox
\end{table}

\section{Violation Rates Across All Criteria}
\label{app:violations}

Table~\ref{tab:full_violations} reports aggregate and per-document
transitivity violation rates for all four judges across all four criteria.
\crit{Fluency} consistently exhibits the highest fraction of documents
with at least one violation.

\begin{table}[H]
\centering
\small
\begin{tabular}{llcc}
\toprule
\textbf{Judge} & \textbf{Criterion} & \textbf{Agg.\ $\bar{\rho}$} & \textbf{\% docs $\geq$1} \\
\midrule
\multirow{4}{*}{\judge{LLaMA-3.1-70B}}
  & Coherence   & 0.008 & 33.3\% \\
  & Consistency & 0.012 & 36.7\% \\
  & Fluency     & 0.033 & \textbf{66.7\%} \\
  & Relevance   & 0.011 & 36.7\% \\
\midrule
\multirow{4}{*}{\judge{Qwen-2.5-72B}}
  & Coherence   & 0.022 & 50.0\% \\
  & Consistency & 0.025 & 56.7\% \\
  & Fluency     & 0.024 & 60.0\% \\
  & Relevance   & 0.023 & 50.0\% \\
\midrule
\multirow{4}{*}{\judge{Mistral-Small}}
  & Coherence   & 0.041 & 50.0\% \\
  & Consistency & 0.036 & 63.3\% \\
  & Fluency     & 0.034 & 53.3\% \\
  & Relevance   & 0.029 & 53.3\% \\
\midrule
\multirow{4}{*}{\judge{GPT-4o-mini}}
  & Coherence   & 0.014 & 46.7\% \\
  & Consistency & 0.013 & 43.3\% \\
  & Fluency     & 0.026 & 53.3\% \\
  & Relevance   & 0.020 & 53.3\% \\
\bottomrule
\end{tabular}
\caption{Aggregate and per-document violation rates across all four
criteria. \crit{Fluency} consistently shows the highest fraction of
documents with $\geq 1$ violation, reaching $66.7\%$ for
\judge{LLaMA-3.1-70B}.}
\label{tab:full_violations}
\end{table}

\section{Coverage vs.\ \texorpdfstring{$\alpha$}{Alpha} Curves}
\label{app:coverage}

\begin{figure}[H]
  \centering
  \includegraphics[width=\linewidth]{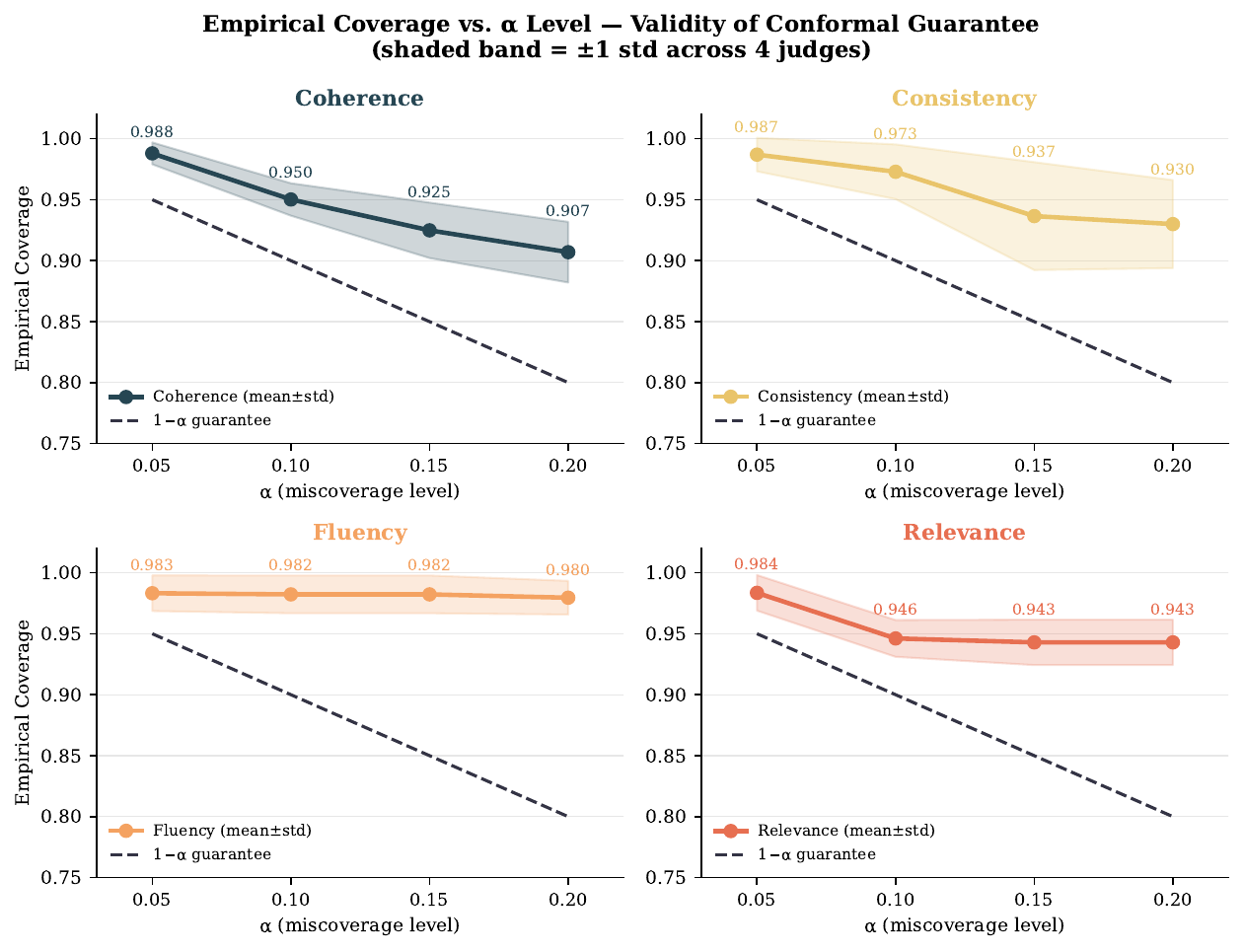}
  \caption{%
    \textbf{Empirical coverage vs.\ $\alpha$.}
    Shaded bands: $\pm1$ std across the four judges.
    Dashed line: theoretical guarantee $1{-}\alpha$.
    Coverage meets or exceeds the guarantee at every operating
    point for all four criteria.}
  \label{fig:coverage}
\end{figure}

\end{document}